\documentclass[conference]{IEEEtran}
\IEEEoverridecommandlockouts
\usepackage{paralist, tabularx}
\usepackage{graphicx}
\usepackage[dvipsnames,table,x11names]{xcolor}
\usepackage{hyperref}
\usepackage{booktabs} 
\usepackage{multirow}
\usepackage{makecell}
\usepackage{siunitx} 
\usepackage{enumitem}

\usepackage{comment} 
\usepackage{marginnote}
\usepackage{amsmath}
\usepackage{amssymb}
\usepackage{amsfonts}

\newcommand{\mypar}[1]{\vspace{1pt}\noindent\textbf{#1.}}
\newcommand{\mypartwo}[1]{\noindent\textit{#1.}}

\begin{document}

\title{LLMs that Understand Processes: Instruction-tuning for Semantics-Aware Process Mining}

\author{\IEEEauthorblockN{Vira Pyrih}
\IEEEauthorblockA{\textit{Faculty of Computer Science} \\
\textit{University of Vienna}\\
Vienna, Austria \\
a12228590@unet.univie.ac.at}
\and
\IEEEauthorblockN{Adrian Rebmann}
\IEEEauthorblockA{\textit{SAP Signavio}\\
Berlin, Germany \\
adrian.rebmann@sap.com}
\and
\IEEEauthorblockN{Han van der Aa}
\IEEEauthorblockA{\textit{Faculty of Computer Science} \\
\textit{University of Vienna}\\
Vienna, Austria \\
han.van.der.aa@univie.ac.at}
}

\maketitle

\begin{abstract}
Process mining is increasingly using textual information associated with events to tackle tasks such as anomaly detection and process discovery. Such semantics-aware process mining focuses on what behavior should be possible in a process (i.e., expectations), thus providing an important complement to traditional, frequency-based techniques that focus on recorded behavior (i.e., reality).
Large Language Models (LLMs) provide a powerful means for tackling semantics-aware tasks. However, the best performance is so far achieved through task-specific fine-tuning, which is computationally intensive and results in models that can only handle one specific task. 
To overcome this lack of generalization, we use this paper to investigate the potential of instruction-tuning for semantics-aware process mining.
The idea of instruction-tuning here is to expose an LLM to prompt-answer pairs for different tasks, e.g., anomaly detection and next-activity prediction, making it more familiar with process mining, thus allowing it to also perform better at unseen tasks, such as process discovery.
Our findings demonstrate a varied impact of instruction-tuning: while performance considerably improved on process discovery and prediction tasks, it varies across models on anomaly detection tasks, highlighting that the selection of tasks for instruction-tuning is critical to achieving desired outcomes.
\end{abstract}


\section{Introduction}

Process mining leverages event data to analyze and improve the execution of business processes, supporting tasks such as process discovery, anomaly detection, and next-activity prediction. Traditionally, process mining techniques have relied heavily on frequency-based methods, focusing on how often certain sequences of activities occur in event logs. Recently, a growing body of research has begun to explore semantics-aware process mining, which incorporates the meaning of activities---often expressed through textual labels—into analysis~\cite{van2021natural, caspary2023does, norouzifar2024bridging}. This shift is driven by advances in natural language processing (NLP), particularly the emergence of large language models (LLMs) with strong capabilities in understanding and generating human language.

Despite the promise of semantics-aware process mining and initial work using fine-tuned LLMs for individual tasks~\cite{caspary2023does,busch2024xsemad,rebmann2025onthepotential}, such as anomaly detection and process discovery, current approaches remain limited by their lack of generalization. Fine-tuning requires separate models for each task and format, leading to inflexible solutions that cannot be used across tasks.

These issues can be overcome through \textit{instruction-tuning}, which is a method for fine-tuning LLMs using 
exemplary pairs of prompts and desired outputs~\cite{wei2021finetuned}, covering different NLP tasks. This enables LLMs to generalize across tasks, by aligning their internal representations with the meaning of the instructions, rather than solely relying on patterns seen in task-specific fine-tuning.
As a result, instruction-tuned models can adapt to tasks and prompts not seen during training.
Although instruction-tuning has shown success in other domains~\cite{zhang2023instruction}, its effectiveness has not yet been explored in process mining. 

Therefore,  we present the first systematic investigation of instruction-tuned LLMs for semantics-aware process mining, aiming to answer the question: \textit{Can instruction-tuning be used to improve the performance of LLMs on unseen process mining tasks?} 
Essentially, we thus assess whether exposing an LLM to certain process mining tasks, such as anomaly detection and next-activity prediction, helps it learn underlying behavioral relations among activities, which can then be leveraged for other tasks, like process discovery.
To operationalize this, we build on previous work that introduced benchmark tasks and datasets for evaluating LLMs in this domain~\cite{rebmann2025onthepotential}. 
In this manner, we evaluate the ability of instruction-tuned LLMs to generalize across various, control-flow-oriented process mining tasks, including classification tasks such as anomaly detection and next-activity prediction, and generative tasks in the form of process discovery (for both DFGs and process trees).
{Our experiments compare models that have been instruction-tuned to both untuned and task-specific models.}

{Our findings suggest that instruction-tuning is a promising path toward more scalable applications of LLMs for semantics-aware process mining, particularly for discovery and predictive tasks. For these tasks, instruction-tuned models exhibit improved generalization capabilities, but our results also reveal challenges in generalization for anomaly detection.}
{The main contribution of this work is the first empirical evidence that a single, instruction-tuned LLM can generalize across multiple semantics-aware process mining tasks, challenging the paradigm of single-task fine-tuning. This has the practical implication of enabling more flexible and scalable process analysis tools, while also defining the research challenge of improving generalization for classification-oriented tasks.}

The remainder of this paper is organized as follows: \autoref{sec:preliminaries} introduces key concepts and \autoref{sec:tasks} outlines the targeted process mining tasks. \autoref{sec:instructiontuning} details our instruction-tuning approach for LLMs. \autoref{sec:setup} and \autoref{sec:results} describe the experimental setup and results, respectively. We discuss related work in \autoref{sec:relatedwork} and conclude in \autoref{sec:conclusion}.

\section{Preliminaries}
\label{sec:preliminaries}

This section introduces preliminaries used in the remainder.

\mypar{Event Data} We adopt a simple event model that focuses on the control-flow of a process. A trace $\sigma = \langle a_1, \dots, a_n \rangle$ captures a single process instance as a sequence of activities $a_i \in \mathcal{A}$, where $\mathcal{A}$ is the universe of possible activities.
An event log $L$ is a finite multi-set of traces. $A_L \subset \mathcal{A}$ denotes the set of activities that appear in the traces of $L$.

\mypar{Directly-Follows Graphs}
A directly-follows graph (DFG) captures which activities in a process can (or have been observed to) directly succeed each other. We define a DFG as a tuple $D = (A, F)$, with $A$ as the DFG's nodes and $F$ as a set of ordered pairs corresponding to its edges. Each edge $(x,y)\in F$ {indicates} that an activity $x \in A$ can be directly followed by activity $y \in A$, which we also denote as $x > y$.

\mypar{Process Trees}
A process tree is a hierarchical representation of a process, using activities and a set of operators $O = {\rightarrow, \times, \wedge, \circlearrowleft}$, denoting sequence, choice, parallelism, and looping. Leaves are activities (or silent activity $\tau$); internal nodes define behavior recursively~\cite{van2022foundations}.

\mypar{Eventually-Follows Relations} 
An eventually-follows relation is a more relaxed ordering relation than the directly-follows relation used in DFGs, focusing on activities that can either directly or indirectly follow each other. 
For a trace $\sigma = \langle a_1,...,a_n\rangle$, any activity pair $a_i, a_j$ with $1 \leq i < j \leq n$ is said to be in an eventually-follows relation, denoted as $a_i \prec a_j$.

\section{Semantics-aware Process Mining Tasks}
\label{sec:tasks}
Our work builds on five semantics-aware
mining tasks introduced in earlier work~\cite{rebmann2025onthepotential}. These are designed to assess the ability of language models to reason about control-flow behavior based solely on the semantics of activities, without access to historical event data. They are defined as follows:

\mypar{Trace-Level Semantic Anomaly Detection (T-SAD)}
    Semantic anomaly detection involves assessing whether observed process behavior makes logical sense or not. In this regard, T-SAD is a classification task where a trace $\sigma$, provided along with a set of possible process activities $A \subseteq \mathcal{A}$, must be classified as being either semantically correct or anomalous. 
    For instance, trace $\sigma = \langle$\emph{register application}, \emph{approve application}, \emph{review application}$\rangle$ should be classified as anomalous, since an application should be reviewed before it is approved.
    Note that including the set $A$ as input enables the identification of anomalies involving missing activities.

\mypar{Activity-Level Semantic Anomaly Detection (A-SAD)}
    A-SAD is a more fine-granular task, where,  given a trace $\sigma = \langle a_1, \ldots, a_n \rangle$ and a set of possible activities $A$, each eventually-follows relation 
$a_i \succ a_j$, with $1 \leq i < j \leq n$, should be classified as semantically correct or anomalous. 
    For instance, for the aforementioned trace, the relation \emph{register application} $\succ$ \emph{approve application} should be recognized as valid and \emph{approve application} $\succ$ \emph{review application} as anomalous.

\mypar{Semantic Next-Activity Prediction (S-NAP)} 
    S-NAP is a classification task in which, given a 
    incomplete trace $\sigma = \langle a_1, \ldots, a_n \rangle$, the task is to select the most suitable activity $a_{n+1}$ from a set of possible activities $A$.
    For instance, given $\sigma = \langle$\emph{create PO}, \emph{approve PO}$\rangle$ and $A=\{$\emph{create PO}, \emph{approve PO}, \emph{create invoice}, \emph{make payment}$\}$, the next activity should be \emph{create invoice}.

\mypar{Semantic Directly-Follows Graph Discovery (S-DFD)}  
  S-DFD is a generation task\footnote{This is a generation task because an answer must be generated, rather than selected from a set of options, as done for the classification tasks.} in which, given a set of activities $A$, the goal is to produce a graph  $D = (A, F)$, where $F$ contains all valid directly-follows relations between activities in $A$.
For instance, given a set of activities $\{$\emph{create PO}, \emph{approve PO}, \emph{reject PO}, \emph{create invoice}$\}$, the semantic DFG should include edges such as (\emph{create PO}, \emph{approve PO}) and (\emph{create PO}, \emph{reject PO}), whereas it should not include (\emph{reject PO}, \emph{create invoice}).

\mypar{Semantic Process Tree Discovery (S-PTD)}  
S-PTD is a more complex discovery task than S-DFD. Specifically, given a set of activities $A$, the goal is to generate a process tree whose structure reflects the behavioral constraints in a process, such as parallelism, choices, and sequential behavior.
For instance, given the set $\{$\emph{create PO}, \emph{approve PO}, \emph{reject PO}, \emph{create invoice}$\}$, the generated tree should be equivalent to
$\rightarrow$(create PO, $\times$(reject PO, $\rightarrow$(approve PO, create invoice))), capturing that, after creating a PO, it can be either rejected or approved, where the latter case leads to an invoice being created.

\section{Instruction-tuning for Semantics-Aware Process Mining}
\label{sec:instructiontuning}
In this section, we explain how to specialize LLMs for semantics-aware process mining tasks using instruction-tuning, also contrasting it to the in-context learning and fine-tuning methods that have been used in previous works. 
We begin by outlining these different specialization strategies (\autoref{sec:instructiontuning:strategies}), before describing the development of our process mining instruction dataset (\autoref{sec:instructiontuning:dataset}).

\subsection{Specializing LLMs for Semantics-Aware Process Mining}
\label{sec:instructiontuning:strategies}
\begin{figure}[h!]
    \centering
    \includegraphics[width=.85\linewidth]{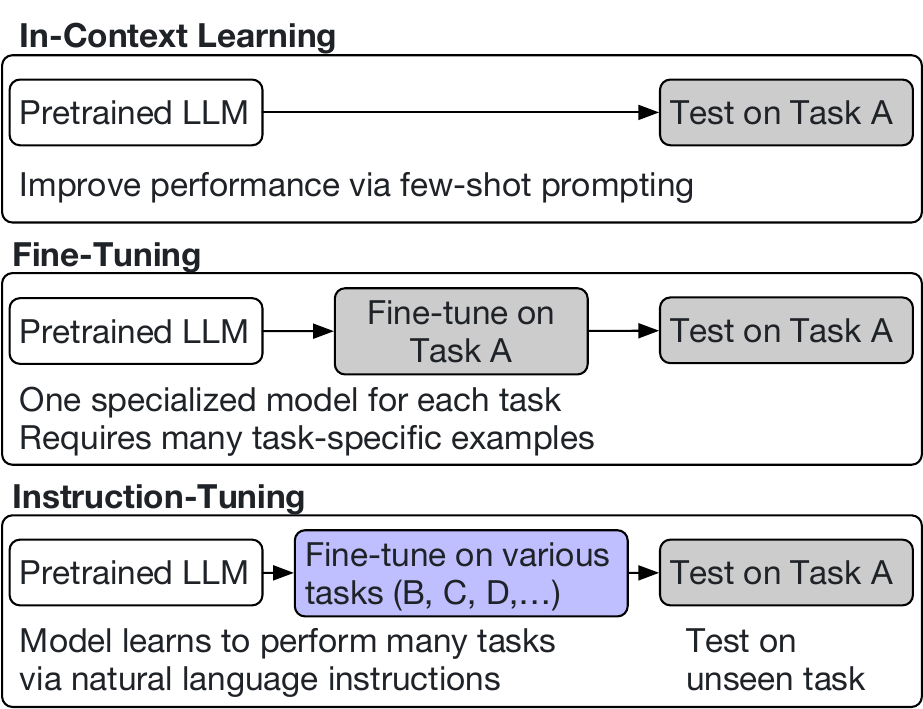}
    \caption{Comparison between in-context learning, fine-tuning, and instruction-tuning. Adapted from Wei et al.~\cite{wei2021finetuned}}
    \label{fig:specialization_strategies}
\end{figure}

Neural language models based on the Transformer architecture~\cite{vaswani_attention_2017, devlin2019bert} generally fall into two categories. Bidirectional models, often called encoders, are typically pretrained using masked language modeling, where they predict masked tokens using context from both directions. 
Unidirectional models, known as decoders, are trained with autoregressive language modeling, predicting the next token based solely on the preceding ones.
LLMs represent large instances of these decoder models, with billions of parameters. Their initial development involves large-scale autoregressive pretraining, which equips them with broad natural language understanding capabilities.
After initial pretraining, LLMs can be specialized for specific tasks. Three common strategies for this are illustrated in \autoref{fig:specialization_strategies} and described next.

\mypar{In-Context Learning}
ICL offers a mechanism to elicit task-specific capabilities from an LLM without modifying its underlying parameters. 
This is done using a prompt that typically includes a natural language description of the task, followed by a few illustrative examples (``shots''), consisting of input-output pairs that demonstrate the desired solution. The actual query instance for which a solution is needed is appended to this context, and the LLM generates the output autoregressively, leveraging the provided examples to infer the task pattern~\cite{dong2022survey}. 
{ICL relies on a model's pretrained knowledge to perform tasks. The model's reasoning is prompted based on a few provided examples.}
For instance, for the A-SAD task, an LLM is asked to classify a set of eventually-follows pairs as anomalous or not, while also providing it with a few examples of valid and invalid pairs.

\mypar{Fine-Tuning}
Fine-tuning adapts a pretrained language model to a specific task by continuing its training on a labeled dataset relevant to that task, updating the model's parameters accordingly.
This is done by providing the model with an extensive set of labeled task instances, such as eventually-follows pairs of activities along with the correct label (\emph{Valid} or \emph{Anomalous}) for A-SAD, yet typically without providing task instructions. 
In this manner, fine-tuning allows models to achieve considerably higher performance on the targeted task compared to ICL~\cite{rebmann2025onthepotential}. However, a clear downside of this strategy is that obtained models can only be used for the specific task they were fine-tuned on. 

\mypar{Instruction-Tuning}
Instruction-tuning enables LLMs to follow natural language instructions and provide solutions to described tasks. Typically, LLMs are already instruction-tuned following their initial autoregressive pretraining, using datasets containing a wide array of common NLP tasks that are described via natural language instructions~\cite{zhang2023instruction} . This step aims to improve generalization across diverse capabilities, covering tasks such as text classification, summarization, question answering, information extraction, and text rewriting~\cite{wang2022super}.

While such general instruction-tuning provides broad instruction-following competence, domain-specific instruction-tuning represents a more focused specialization strategy. 
This involves further training the model on a curated dataset comprising instructions for solving tasks that are relevant to the domain, which, in our case, is process analysis. 
Domain-specific tasks thus include, e.g., ``Classify this trace as anomalous or valid'' and ``Given this partial trace and list of possible activities, predict the next one'', paired with corresponding domain-specific inputs such as traces or activity lists and the desired outputs. 
Unlike single-task fine-tuning, which primarily optimizes performance for a fixed input-output format, domain-specific instruction-tuning focuses on enhancing the model's ability to interpret the intent behind various instructions phrased using process mining terminology and concepts. The goal is to improve the LLM's zero-shot or few-shot generalization capabilities across different types of process analysis tasks, making it more versatile compared to models relying solely on generic instruction following learned during initial training or adapted via ICL.

\subsection{Creating a Process Mining Instruction Dataset}
\label{sec:instructiontuning:dataset}
A prerequisite for instruction-tuning is a dataset of labeled instruction-task instances. This section describes how we established such a dataset for semantics-aware process mining tasks, which is publicly available~\cite{pyrih_2025_15498373}.

\mypar{Labeled-instance Datasets}
As a basis for generating our instruction dataset, we use five datasets from our earlier work~\cite{rebmann2025onthepotential}, each consisting of labeled instances of one of the semantics-aware process mining tasks (\autoref{sec:tasks}). 
They were derived from a collection of over 15,000 process models from the SAP-SAM collection~\cite{sola2023sap}, covering a wide range of process types and domains.\footnote{For details on their establishment and characteristics, we refer to~\cite{rebmann2025onthepotential}.}

To avoid the inclusion of duplicate task instances, we cleaned these available datasets using task-specific rules:
\begin{itemize}[topsep=0pt,leftmargin=1em]
\item T-SAD: Filtered for unique (activity, full trace) combinations (184,004 instances).
\item A-SAD: Filtered for unique (activity, activity pair) combinations (316,308 instances).
\item S-NAP: Removed completed traces and duplicate prefixes for the same next activity, preserving cases with multiple valid outcomes (575,339 instances).
\item S-DFD/S-PTD: Filtered for unique input activity sets (15,580 instances).
\end{itemize}

\mypar{Creating Instruction-Task Instances}
Since the labeled-instance datasets only include input and output pairs $(t, o)$ (e.g., for T-SAD, $t$ is a trace with its corresponding output label $o$ as \textit{Valid} or \textit{Anomalous)}, rather than prompts, we create instruction-task instances by enriching the available instances with natural language instructions to be used by an LLM.

To this end, we follow the \emph{human-oriented instructions} approach~\cite{lou2024large}, which aims to establish instructions that are understandable by non-expert users and typically take the form of descriptive, paragraph-style text. They include a brief statement that frames the identity of the model, contextual information such as task descriptions, definitions, and specific instructions, which has been shown to enhance generalization particularly well~\cite{lou2024large}. 

\begin{figure}[h!]
    \centering
    \includegraphics[width=.9\linewidth]{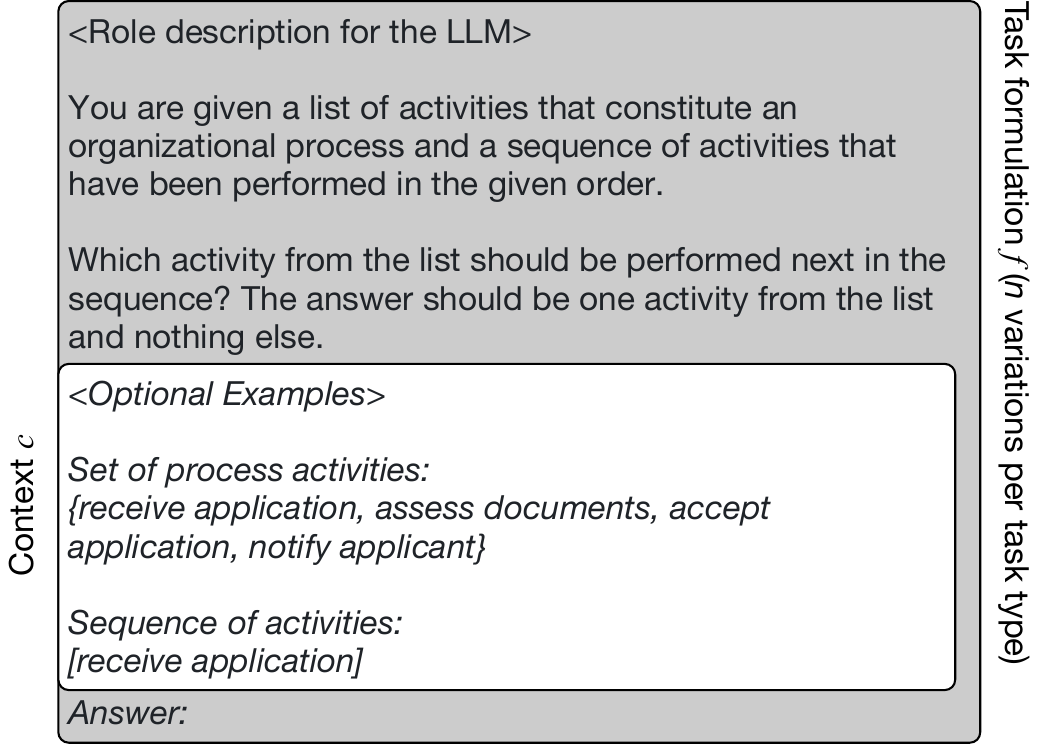}
    \caption{Instruction-task-instance for the S-NAP task.}
    \label{fig:snap_normal}
\end{figure}

In line with the state of the art~\cite{wang2022super, longpre2023flan}, each labeled instruction-task instance is a pair $\phi=(i, o)$ consisting of an instruction $i$, which we detail below, and an expected output $o$, which we directly derive from the output label $o$ of the original task instance (reformatted as a natural language response). As illustrated in~\autoref{fig:snap_normal} for S-NAP, an instruction-task instance is a tuple $i = (f, c)$. The task formulation $f$ is a natural language description of the task, including the role description of the LLM, requirements for the output format, and a closing phrase prompting the model to generate the desired result. The context $c$ is a textual representation of the task instance (e.g., a trace or activity set) and any additional contextual information (e.g., examples) that the model must process.

\noindent To increase the diversity and robustness of the resulting instruction dataset, we adhere to the following best practices~\cite{zhang2023instruction}:

\mypartwo{Varying task formulations} We establish six formulations for each task. This exposes the model to different phrasings of the same underlying problem, thereby improving its generalization capabilities to new, unseen instructions. 
For instance, task formulations for A-SAD range from a formal instruction to ``Determine whether it is valid for the first activity to occur before the second'', which presents the activities separately and constrains the output to ``Provide either True or False as the answer and nothing else'' to more direct phrasings that embed the activities within the prompt, such as ``Is the order (first: {act1}, second: {act2}) acceptable...?'' while requesting the model to simply ``Answer with True or False.'' The complete set of variations for all tasks is available in our repository.

\mypartwo{Incorporating instructions about negative or undesired behavior} Real-world process mining often involves identifying or reasoning about undesired, incorrect, or incomplete process behavior. To reflect this, we create instructions that focus on such negative aspects where appropriate. Here, \emph{negative} refers to instructions prompting the model to identify, complete, or reason about incorrect, incomplete, or undesirable process elements or sequences.
For example, for S-NAP, a negative task instance might describe an ongoing process execution that is missing an important activity that should have occurred earlier. The model is then tasked to identify this missing activity from any point in the prior sequence leading up to the last observed activity. This helps prevent LLMs from overfitting solely to desired process behavior and ensures they can handle the variance seen in practical applications.

\mypartwo{Inverting task objectives} We generate reversed versions of a given task by flipping its objective, if the task structure allows. This can lead to both \emph{positive} and \emph{negative} inversions.
\begin{itemize}[topsep=0pt,leftmargin=1em]
    \item \textit{Positive inversion} typically involves the model generating or identifying correct, preceding, or completing elements for an assumed valid partial context. For example, for S-NAP, a positive inversion involves providing the model with a known subsequent activity in a trace and asking it to generate a plausible sequence of activities that might have occurred \emph{before} this given activity. For A-SAD, a positive inversion might involve completing a valid partial activity pair by having the model choose an activity that can legitimately follow a given one.

    \item \textit{Negative inversion} involves the model identifying elements that render a task instance incorrect, highlight what should not occur, or complete a scenario in a way that demonstrates an invalid or undesirable path. 
    For instance, a negative inversion of A-SAD, could involve asking the model to choose an activity that would create an invalid eventually-follows pair if it followed a given activity.
\end{itemize}
For T-SAD and A-SAD, the nature of the inversion depends on whether the original task instance presented is valid or anomalous. For instance, if an A-SAD instance is an anomalous pair, a negative inversion asks to create an activity pair such that it is anomalous, while a positive inversion for a valid pair asks to create a valid pair. For S-NAP, the inversion is chosen to be positive or negative with equal probability. For the discovery tasks, since generating a positively inverted task is not meaningful (e.g., generating a set of activities from a given DFG), we only apply negative inversions.

 Following these best practices, we transform each original task instance into a labeled instruction-task instance $\phi$. We begin by randomly selecting a task formulation $f$ from the available ones. The original task instance $t$ is then transformed into a corresponding context description $c$ (occassionally using inversion). These components, $f$ and $c$, form the instruction pair $i = (f, c)$, which, together with the original output label $o$, constitutes the labeled instruction-task instance $\phi = (i, o)$. In this way, we obtain an instruction data set, made publicly available~\cite{pyrih_2025_15498373}, whose characteristics are shown in \autoref{tab:data_characteristics}.

\begin{table}[!htbp] 
\centering
\setlength{\tabcolsep}{3.5pt} 
\caption{Characteristics of the instruction dataset.}
\label{tab:data_characteristics}
\begin{tabular}{lrccc}
\toprule
\textbf{Task} & \textbf{\makecell{Samples \\ (Total)}} & \textbf{\makecell{Normal \\ (\%)}} & \textbf{\makecell{Neg. Inv. \\ (\%)}} & \textbf{\makecell{Pos. Inv. \\ (\%)}} \\
\midrule
A-SAD & 316,308 & \hphantom{1}80 & \hphantom{1}10 & \hphantom{1}10 \\
T-SAD & 184,004 & \hphantom{1}80 & \hphantom{1}10 & \hphantom{1}10 \\
S-NAP & 575,339 & \hphantom{1}80 & \hphantom{1}10 & \hphantom{1}10 \\
S-DFD & 15,580 & \hphantom{1}80 & \hphantom{1}20 & \hphantom{10}0 \\
S-PTD & 15,580 & 100 & \hphantom{10}0 & \hphantom{10}0 \\
\bottomrule
\end{tabular}
\end{table}

\section{Experimental Setup}
\label{sec:setup}
We evaluate the generalization capabilities of LLMs in solving semantics-aware process mining tasks, primarily examining the difference between models with and without domain-specific instruction-tuning.
To this end, this section details our experimental setup; the code used to conduct the experiments is available in our repository.\footnote{\url{https://github.com/pirogtm7/it4pm}}

\mypar{Task Grouping}
We evaluate the LLMs' ability to generalize across different process mining objectives by grouping the  tasks into three coherent groups: \textit{Anomaly} (A-SAD and T-SAD), \textit{Prediction} (S-NAP), and \textit{Discovery} (S-DFD and S-PTD). 
This enables a leave-one-group-out evaluation setup, where models are trained on two groups (e.g., Anomaly and Prediction) and tested on the held-out group (e.g., Discovery).
By using groups rather than individual tasks, we avoid bias caused by having an anomaly detection (or discovery) task in both the training and the test set.

\mypar{Dataset Splitting}
To ensure reproducibility and consistent comparisons, we construct our leave-one-group-out folds using fixed data splits for each task, adopted from our previous work~\cite{rebmann2025onthepotential}. For each task, instruction instances are derived from a distinct set of underlying process models. These models are partitioned to allocate 70\% for generating training instances, 20\% for validation, and 10\% for testing.
These pre-defined, task-specific splits are then combined to form the datasets for each leave-one-group-out fold as follows:

\mypartwo{Training} For a given fold, the training data is formed by combining the training splits of the tasks belonging to the two in-fold groups.

\mypartwo{Validation} The combined validation splits of tasks in the held-out group guide model checkpoint selection during training.

\mypartwo{Test} The combined test splits of tasks in the held-out group guide the evaluation of final model performance.

\noindent
Building on fixed individual task splits guarantees that, for instance, the test set for the \emph{Discovery} group remains identical, whether used in our leave-one-group-out evaluation or by other researchers for direct comparisons on S-DFD and S-PTD tasks.

\mypar{Training Data Sampling}
To balance task representation for instruction-tuning, we sample training sets using examples-proportional mixing~\cite{raffel2020exploring} from constituent tasks, generally capping contributions at 30,000 samples per task. This cap is adjusted when holding out the \emph{Discovery} group: S-NAP contributes 60,000 samples to balance the 60,000 from two anomaly detection tasks.
The characteristics of the resulting training folds are shown in \autoref{tab:train_sample_distribution}.
\begin{table}[h!]
\centering
\caption{Training sample distribution per test group}
\label{tab:train_sample_distribution} 
\setlength{\tabcolsep}{3pt} 
\begin{tabular}{@{}cc r r ccc@{}} 
\toprule
\textbf{Test} & \textbf{Training} & \multirow{2}{*}{\textbf{Samples}} & \multirow{2}{*}{\textbf{Share}} & \multicolumn{3}{c}{\textbf{Prompt Types (\%)}} \\
\textbf{Group}& \textbf{Task} &  & &  {Norm.} & {Neg.Inv.} & {Pos.Inv.} \\
\midrule
\multirow{4}{*}{Anomaly}
& S-NAP & 30,000 & 49.05 & 31.39 & 9.81 & 7.85 \\
& S-DFD & 15,580 & 25.47 & 20.38 & 5.09 & - \\
& S-PTD & 15,580 & 25.47 & 25.47  & - & - \\
& \textbf{Total} & \textbf{61,160} & \textbf{100.0} & \textbf{77.25} & \textbf{14.90} & \textbf{7.85} \\
\midrule
\multirow{5}{*}{Prediction}
& A-SAD & 30,000 & 32.91 & 26.33 & 3.24 & 3.35 \\
& T-SAD & 30,000 & 32.91 & 26.33 & 3.29 & 3.29 \\
& S-DFD & 15,580 & 17.09 & 17.09 & - & - \\
& S-PTD & 15,580 & 17.09 & 17.09 & - & - \\
& \textbf{Total} & \textbf{91,160} & \textbf{100.00} & \textbf{86.84} & \textbf{6.53} & \textbf{6.64} \\
\midrule
\multirow{4}{*}{Discovery}
& A-SAD & 30,000 & 25.00 & 20.00 & 2.46 & 2.54 \\
& T-SAD & 30,000 & 25.00 & 20.00 & 2.50 & 2.50 \\
& S-NAP & 60,000 & 50.00 & 40.00 & - & 10.00 \\
& \textbf{Total} & \textbf{120,000} & \textbf{100.00} & \textbf{80.00} & \textbf{4.96} & \textbf{15.04} \\
\bottomrule
\end{tabular}
\vspace{-1em}
\end{table}

\mypar{Large Language Models}
For our instruction-tuning experiments, we selected two powerful decoder language models: Llama 3 70B Instruct\footnote{\url{https://huggingface.co/unsloth/Llama-3.3-70B-Instruct-bnb-4bit}} and Mistral Large 2 Instruct (v.\ 2407)\footnote{\url{https://huggingface.co/unsloth/Mistral-Large-Instruct-2407-bnb-4bit}}. These were chosen for their robust instruction-following capabilities, model size, and open-source availability. 

\mypar{Instruction-Tuning Procedure}
We use parameter-efficient fine-tuning based on quantized low-rank adaptation (LoRA)~\cite{hu2021lora}, using pre-quantized 4-bit models, configuring LoRA with rank  $r = 16$ and scaling factor $\alpha = 16$ to balance model expressiveness while minimizing the risk of overfitting. We verified that higher ranks did not lead to consistent improvements. We set the learning rate to $1\times10^{-5}$ and adjust it dynamically using a linear scheduler. We perform training with a batch size of 8 per GPU and 4 gradient accumulation steps, resulting in an effective batch size of 32\footnote{For Mistral's prediction and anomaly tasks, the batch size was reduced to 4 (effective size 16) due to memory constraints.}. We fine-tune for 3 epochs, using a maximum sequence length of 1024 tokens. We use the AdamW optimizer with a weight decay of 0.01 to improve regularization. 

During training, we use standard cross-entropy as the loss-function and monitor validation performance every 250 steps using the combined validation splits of tasks in the held-out group. For model selection, instead of relying solely on cross-entropy loss, we compute task-specific custom metrics that are better aligned with the target tasks. When validating multiple tasks simultaneously (for the Anomaly and Discovery groups), we average the metric improvements across tasks and select the checkpoint achieving the highest overall improvement.

\mypar{Performance Measures}
We assess the obtained results using established performance measures:

\mypartwo{Anomaly Detection and Prediction Performance}
Since anomaly detection (T-SAD and A-SAD) and next-activity prediction (S-NAP) involve classification, we assess them using the macro F$_1$-score. This measure ensures that all classes of the respective task contribute equally to the overall performance, irrespective of their size. The macro F$_1$-score is the simple average of the F$_1$-scores for each individual class (e.g., valid and anomalous traces in T-SAD), where the F$_1$-score for a class is the harmonic mean of its precision and recall.

\mypartwo{Discovery Performance}
For the discovery tasks (S-DFD and S-PTD), we measure performance through footprint-based fitness~\cite{carmona2018conformance}. This measure facilitates the comparison of sets of allowed execution sequences by using pairwise behavioral relations. As such, it considers whether a discovered model allows for the same behavior as the ground-truth model.

\section{Experimental Results}
\label{sec:results}
We report on the LLMs' performance on the five tasks, comparing their performance with and without instruction-tuning, followed by an in-depth analysis of the results.

\subsection{Main Results}
\label{sec:main_results}

Table~\ref{tab:main_results} reports on the overall results.\footnote{{These results should be considered in light of Mistral's considerably longer tuning times due to its larger architecture (123B vs. 70B parameters). For instance, an epoch for the \emph{Discovery} test group took 37h vs 21h for Llama.}} We observe mixed impacts of instruction-tuning on performance. While the instruction-tuned models consistently outperform the base models for the prediction (S-NAP) and discovery (S-DFD and S-PTD) tasks, the effect on anomaly detection tasks (A-SAD and T-SAD) differs between models.

\begin{table}[h!]
\centering
\caption{Performance across models and tasks.}
\label{tab:main_results}
\sisetup{detect-weight=true, detect-mode=true}
\begin{tabular}{@{} l S[table-format=1.3] S[table-format=1.3] S[table-format=1.3] S[table-format=1.3] @{}}
\toprule
{\textbf{Task (Metric)}} & {\textbf{Llama}} & {\textbf{Llama}} & {\textbf{Mistral}} & {\textbf{Mistral}} \\
{}                       & {\textbf{Base}}  & {\textbf{IT}}    & {\textbf{Base}}    & {\textbf{IT}}      \\
\midrule
A-SAD (macro F$_1$) & 0.594 & 0.562 & 0.421 & \textbf{0.679} \\
T-SAD (macro F$_1$) & 0.558 & 0.480 & 0.347 & \textbf{0.620} \\
\midrule
S-NAP (macro F$_1$) & 0.525          & \textbf{0.651} & 0.624 & \textbf{0.868} \\
\midrule
S-DFD (Fitness)     & 0.630          & \textbf{0.714} & 0.658          & \textbf{0.770} \\
S-PTD (Fitness)     & 0.621          & \textbf{0.697} & 0.649          & \textbf{0.763} \\
\bottomrule
\end{tabular}
\end{table}

{Specifically, on anomaly detection, Llama IT's performance decreases (T-SAD F$_1$: 0.558 to 0.480), whereas Mistral IT's improves substantially (T-SAD F$_1$: 0.347 to 0.620). {For example, given the sequence of activities $\langle$\emph{Approve application}, \emph{Apply for trip}, \emph{Buy transport tickets}, \emph{Book accommodation}, \emph{Archive trip documents}$\rangle$, the Llama Base correctly identifies the trace as invalid, whereas the Llama IT model incorrectly flags it as valid.} 
This suggests that  instruction-tuning on discovery and predictive tasks may have overwritten some of Llama's semantic anomaly capabilities. However, this effect is not universal, as Mistral IT consistently improves performance compared to its base version.
}
Conversely, for S-NAP, instruction-tuning boosts performance for both models. Llama IT's F$_1$ score increases from 0.525 to 0.651. 
{For instance, given the prefix $\langle$\emph{Check stock availability}, \emph{Retrieve product}$\rangle$, Llama Base suggests \emph{Emit invoice}, whereas Llama IT, more logically, predicts \emph{Confirm order}, i.e., that an order must be confirmed before an invoice is issued.}

Moving to process discovery, instruction-tuning shows substantial benefits for both tasks. For Llama, the fitness score improves from 0.630 (Base) to 0.714 (IT) for S-DFD. Mistral exhibits even larger gains, with fitness increasing from 0.658 (Base) to 0.770 (IT). Similarly, for the S-PTD task, instruction-tuning leads to considerable improvements. Llama's fitness score rises from 0.621 to 0.697 after instruction-tuning. Mistral sees its fitness score increase from 0.649 to 0.763.

Overall, the results reveal a varied impact of instruction-tuning. While Llama IT shows decreased macro F$_1$ scores on A-SAD and T-SAD, Mistral IT shows a considerable improvement. Both models, however, achieve notable improvements on S-NAP. For tasks requiring structured output generation, namely S-DFD and S-PTD, instruction-tuning consistently yields substantial gains in fitness scores for both Llama and Mistral. These improvements confirm that both models considerably benefit from domain-specific instruction-tuning in these scenarios. The enhancement suggests that instruction-tuning not only facilitates task-specific adaptation for structured output but also helps models grasp higher-level process representations, particularly where multiple plausible outputs exist. Notably, Mistral IT consistently outperforms Llama IT on these tasks, as expected from its larger size.
{For instance, Llama IT incorrectly placed \emph{Analyze claim} in sequence after $\times$(\emph{Handle easy claim}, \emph{Handle complex claim}), whereas Mistral IT correctly placed it before, resulting in higher fitness.}

\subsection{In-Depth Analysis}
\label{sec:in_depth_analysis}

Beyond the aggregate performance metrics (Section~\ref{sec:main_results}), we report on qualitative results of instruction-tuned models and their base counterparts, providing more detailed insights into their understanding of process behavior.

\mypar{Improved Instruction Adherence}
We find that instruction-tuned models consistently show improved adherence to specified instruction formats and produce cleaner, more direct outputs compared to base models. 
For instance, in the T-SAD task, Llama IT provided clear \emph{True} or \emph{False} responses, whereas Llama Base often returned ambiguous or irrelevant text. 
Similarly, for S-NAP, Llama IT generated concise next-activity predictions, a considerable improvement over the base model's tendency to generate boilerplate text or code-like structures. 
This enhanced clarity also extends to the discovery tasks: the instruction-tuned models produced better-formatted directly-follows pairs (S-DFD) and more structurally coherent process trees (S-PTD), largely avoiding malformed or incomplete outputs that are common with base models. 
This suggests that instruction-tuning effectively guides the models' focus on the core process analysis task and desired output structure.

\mypar{Superiority in Semantic Discovery Tasks}
The benefits of instruction-tuning are particularly visible in semantic discovery tasks (S-DFD and S-PTD), which require a more holistic understanding of the underlying process and activity relationships. 
While the base models capture basic relations, instruction-tuned models demonstrate a more developed grasp of higher-level process representations. 
For example, in S-PTD, Mistral IT not only achieved higher fitness but also exhibited a qualitatively better interpretation of control-flow operators such as exclusion ($\times$) and sequence ($\rightarrow$) compared to its base version, leading to more semantically sound process trees. 
This suggests that exposure to diverse process-related instructions helps instruction-tuned models develop a better internal representation of process structures.

\mypar{Challenges in Classification-Oriented Tasks}
For classification tasks such as anomaly detection (A-SAD and T-SAD), the impact of instruction-tuning was model-dependent. For Llama IT, the impact was unfavorable, with the model underperforming its base counterpart. In contrast, Mistral IT showed considerable performance gains on these same tasks.
A reason for Llama's less favorable performance lies in a prediction bias for A-SAD and T-SAD towards predicting \emph{True} (valid) for the instruction-tuned model, which Mistral could compensate, likely due to its larger size.
For T-SAD, Llama's bias became particularly clear, with 89.1\% of predictions being \emph{True}, while the test data has an even label distribution. This indicates that while the IT model improves instruction adherence, it does not resolve issues such as label bias if not carefully managed in the instruction dataset design and task formulation. 
The difficulty of these tasks, requiring an inference based solely on activity semantics without context information, likely contributes to these challenges. {Investigating how to address them remains an important direction for future research.}

\mypar{Domain-Specific Differences}
To assess model performance across process domains, we categorized the instruction source process models by industry.\footnote{The detailed categorization procedure and quantitative results are available in our repository.}
Both base and instruction-tuned models exhibit varying performance across different process industries. Instruction-tuned models generally performed well in domains such as \textit{Logistics} and \emph{Education}, which are often characterized by more structured, procedural, and clearly defined processes (e.g., shipment handling, course registration) aligning well with patterns that are learnable by LLMs.
Conversely, domains such as \textit{Healthcare} and \textit{IT/Software} present greater challenges. These areas are characterized by flexible workflows, high variability across cases (e.g., patient treatment, software development), and frequent ad-hoc decisions, making it more difficult for LLMs to generalize behavioral patterns. This underscores that while instruction-tuning enhances adaptability, inherent domain complexity and variability remain substantial challenges.

In summary, instruction-tuning is a promising path to more flexible semantics-aware process mining. It particularly enhances performance on semantic discovery and instruction adherence. Challenges persist in classification and highly variable domains, indicating a need for task refinement and additional model context.

\subsection{Comparison to Fine-Tuning}
{To put the results of our instruction-tuning experiments into context, we compare them with the task-specific fine-tuned (FT) smaller language models from our previous work~\cite{rebmann2025onthepotential}, i.e., the small encoder model RoBERTa\footnote{Note that encoder models do not have text completion capabilities, thus RoBERTa cannot be used for the discovery tasks.} and the Mistral 7B and Llama 8B decoder LLMs.

\begin{table}[h!]
    \centering
    \caption{Performance of Fine-Tuned (FT) LLMs.}
    \label{tab:results:combined_ft}
    \begin{tabular}{@{}lccc@{}}
        \toprule
        \textbf{Task (Metric)} & \textbf{RoBERTa} & \textbf{Mistral 7B} & \textbf{Llama 8B} \\
        \midrule
        \quad T-SAD (macro $F_1$) & 0.77 & 0.79 & 0.79 \\
        \quad A-SAD (macro $F_1$) & 0.85 & 0.88 & 0.88 \\
        \midrule
        \quad S-NAP (macro $F_1$) & 0.63 & 0.68 & 0.69 \\
        \midrule
        \quad S-DFD (Fitness) & - & 0.81 & 0.80 \\
        \quad S-PTD (Fitness) & - & 0.84 & 0.83 \\
        \bottomrule
    \end{tabular}
    \vspace{-1em}
\end{table}

Comparing IT results (\autoref{tab:main_results}) with FT results (\autoref{tab:results:combined_ft}) shows FT models achieve consistently higher scores. For instance, the FT Llama~8B scores 0.88 on the A-SAD task, far surpassing the IT Llama~70B's 0.562. This gap persists even for the much smaller RoBERTa model (scoring 0.85). The smaller fine-tuned LLMs also outperform their 70B+ IT counterparts on both discovery tasks. These results are unsurprising, as FT models are trained on large amounts of in-task data, whereas the IT models have observed none.

However, this superior performance has a substantial drawback: each FT model is a non-generalizing specialist. This requires creating extensive labeled datasets for every task and introduces considerable operational challenges in deploying, maintaining, and updating these models. Furthermore, for classification tasks, the performance gains of fine-tuning large decoder models over encoders are often marginal. As shown in \autoref{tab:results:combined_ft}, the RoBERTa encoder achieves scores that the FT LLMs only slightly surpass, suggesting a smaller encoder may be a more efficient choice for these tasks.

In contrast, the key advantage of IT models is their ability to generalize to unseen tasks, the capability evaluated in our leave-one-group-out setup. Moreover, the crucial advantage of the LLM (decoder) architecture is its suitability for instruction-tuning, as encoders cannot be instruction-tuned to handle diverse, unseen tasks in the same way.

Overall, the choice between FT and IT thus depends on the desired outcome: maximizing performance on a single, static task, or building a more versatile process analysis tool.
}

\section{Related Work}
\label{sec:relatedwork}

The advent of LLMs has resulted in a range of works that explore their potential in the broader process analysis domain, including various applications in process mining~\cite{torres2024mapping}. 

Our work focuses on process mining tasks that can be evaluated using gold-standard benchmarks, enabling rigorous methodological evaluation and fine-tuning. 
{Existing work in this category has predominantly addressed semantic anomaly detection~\cite{caspary2023does,guan2025dabl}, next activity prediction~\cite{pasquadibisceglie2024lupin,oved2025snap} and the discovery of declarative process constraints~\cite{busch2024xsemad}, thus aligning with our A-SAD, T-SAD, S-NAP, and S-PTD tasks (though, the latter only to some extent).}

In contrast, a range of studies apply LLMs to tasks lacking gold-standard assessments. While this makes it harder to objectively evaluate model performance, it allows exploration of a broader set of process mining applications. Such works include those focused on identifying bottlenecks or undesired process behaviors~\cite{berti2023abstractions}, abstracting fine-granular events into higher-level ones~\cite{fani2023llms}, incorporating domain knowledge into process discovery~\cite{norouzifar2024bridging}, and providing explanations in prescriptive process monitoring~\cite{kubrakBMND24}.
To address the lack of standardized evaluation in such settings, Berti et al.~\cite{berti2024pm} proposed a benchmark of process mining analysis questions that enables self-evaluation of LLM performance. While promising, the use of self-evaluation has been shown to introduce potential biases~\cite{panickssery2024llm}, thus requiring careful interpretation of results.

\section{Conclusion}
\label{sec:conclusion}

\mypar{Summary} This work presented the first experimental study evaluating the potential of instruction-tuning to enhance the process mining capabilities of LLMs, aiming to overcome the downsides of task-specific fine-tuning. 
{Our results demonstrate that instruction-tuned models outperform their untuned counterparts on several key process mining tasks. However, the observed performance varies across models in anomaly detection highlights the critical need for careful task selection during the instruction-tuning phase.} Notably, the most substantial performance gains were observed in process discovery tasks, which demand a deep understanding of behavioral dependencies within processes. This suggests that exposing LLMs to a diverse set of process mining tasks can significantly improve their process comprehension.

\mypar{Limitations} 
{The main limitation of our study stems from its controlled experimental design. While necessary for rigorous evaluation, this setup does not fully capture the complexity of real-world settings, as our tasks intentionally focused on the control-flow perspective using structured inputs. Consequently, the robustness of our instruction-tuned models remains untested on noisy, real-world event logs that contain incomplete data or require other perspectives like time and resources. Furthermore, our results for instruction-tuning itself reveal a critical challenge: while it is beneficial for generative tasks, it can hinder performance on anomaly detection. This suggests that the selection of instruction-tuning tasks for classification requires further investigation.}

\mypar{Outlook} Future research should further investigate the applicability of instruction-tuned LLMs to other areas of process mining. In particular, it would be valuable to examine whether the benefits of instruction-tuning on well-defined, gold-standard tasks transfer to tasks lacking such benchmarks (cf. \autoref{sec:relatedwork}). This would involve leveraging structured instruction-tuning to improve performance in more subjective or exploratory settings. Additionally, we plan to explore how instruction-tuned LLMs can be integrated with traditional process mining techniques, such as combining frequency-based process discovery with semantic assessments.

\smallskip
\noindent\textit{Open science:}
\emph{Our evaluation scripts, instruction-tuned models, and more detailed evaluation results are available through our project repository: \url{https://github.com/pirogtm7/it4pm}. Our instruction dataset is published separately~\cite{pyrih_2025_15498373}.}

\bibliographystyle{IEEEtran}
\bibliography{IEEEabrv, refs}
\end{document}